\documentclass{article}
\usepackage[utf8]{inputenc}
\usepackage{main}
\usepackage{microtype}
\usepackage{graphicx}
\usepackage{times}
\usepackage{amsmath}
\usepackage{amssymb}
\usepackage{enumitem}
\usepackage{booktabs}
\usepackage{natbib}
\definecolor{mydarkblue}{rgb}{0,0.08,0.45}
\usepackage[colorlinks=true,linkcolor=mydarkblue,citecolor=mydarkblue,filecolor=mydarkblue,urlcolor=mydarkblue]{hyperref}
\usepackage{fancyhdr}

\usepackage{xspace}
\usepackage{gradient-text}
\usepackage{float}
\usepackage{multirow}

\setlist[itemize,1]{leftmargin=10pt}

\usepackage{tikz}
\usepackage{soul}
\usepackage{colortbl}
\usepackage{subcaption}
\usepackage{bbm}
\usepackage{adjustbox}
\usepackage{makecell}
\usepackage[framemethod=TikZ]{mdframed}

\definecolor{mypositive}{RGB}{0, 128, 0}
\definecolor{mynegative}{RGB}{220, 20, 60}
\definecolor{mypositive}{RGB}{24, 103, 173}
\definecolor{mynegative}{RGB}{255, 128, 0}

\newcommand{\eg}{\emph{e.g.,}\xspace}
\newcommand{\ie}{\emph{i.e.,}\xspace}

\definecolor{customblue}{HTML}{1F77B4}
\definecolor{customorange}{HTML}{FF7F0E}
\definecolor{customgreen}{HTML}{2CA02C}
\definecolor{no}{HTML}{BA8E23}
\definecolor{darkgreen}{RGB}{0,100,0}
\definecolor{darkred}{RGB}{139,0,0}

\newcommand{\deltaScore}[1]{%
\begingroup
\ifdim #1pt > 0pt
\textcolor{darkred}{\scriptsize$^{#1}$}%
\else
\textcolor{darkgreen}{\scriptsize$^{#1}$}%
\fi
\endgroup
}

\usepackage[skins,breakable]{tcolorbox}

\newtcolorbox{boxblue}{enhanced,colback=blue!5!white,colframe=blue!75!black,breakable=true}

\usepackage{listings}
\usepackage{wrapfig}

\lstdefinelanguage{json}{
    basicstyle=\scriptsize\ttfamily,
    numbers=none,
    numberstyle=\tiny,
    stepnumber=1,
    numbersep=5pt,
    showstringspaces=false,
    breaklines=true,
    frame=none,
    backgroundcolor=\color{white},
    literate=
     *{:}{{{\color{black}{:}}}}{1}
      {,}{{{\color{black}{,}}}}{1}
      {\{}{{{\color{black}{\{}}}}{1}
      {\}}{{{\color{black}{\}}}}}{1}
      {[}{{{\color{black}{[}}}}{1}
      {]}{{{\color{black}{]}}}}{1},
}

\newcommand{\cmark}{\textcolor{green!50!black}{\checkmark}}
\newcommand{\xmark}{\textcolor{red!75!black}{\texttimes}}

\newcommand{\data}{\textsc{\gradientRGB{SciMDR}{64,64,64}{179,26,26}}}
\newcommand{\eval}{\textsc{\gradientRGB{SciMDR-Eval}{64,64,64}{179,26,26}}\xspace}

\definecolor{YaleBlue}{RGB}{0, 53, 107}
\definecolor{UChiRed}{RGB}{128, 0, 0}
\definecolor{TCSC}{RGB}{1, 126, 199}

\newcommand{\UChi}{\hspace{.1em}^{\textcolor{UChiRed}{\boldsymbol{C}}}}
\newcommand{\Yale}{\hspace{.1em}^{\textcolor{YaleBlue}{\boldsymbol{Y}}}}
\newcommand{\TCS}{\hspace{.1em}^{\textcolor{TCSC}{\boldsymbol{T}}}}

\newcommand{\huggingface}{\raisebox{-1.5pt}{\includegraphics[height=1.05em]{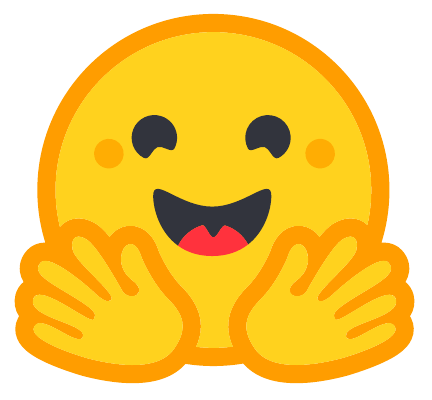}}\xspace}
\newcommand{\github}{\raisebox{-1.5pt}{\includegraphics[height=1.05em]{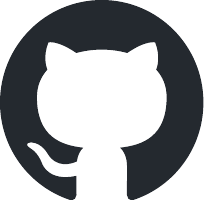}}\xspace}

\begin{document}

\title{\fontsize{15}{17}\selectfont \data: Advancing Scientific Multimodal Document Reasoning}

\author{
\textbf{Ziyu Chen}$\UChi$\thanks{Equal contributions. Correspondence to: Yilun Zhao (\texttt{yilun.zhao@yale.edu})} \qquad
\textbf{Yilun Zhao}$\Yale$\footnotemark[1] \qquad
\textbf{Chengye Wang}$\Yale$ \qquad
\textbf{Rilyn Han}$\Yale$ \\ [3pt]
\textbf{Manasi Patwardhan}$\TCS$ \qquad
\textbf{Arman Cohan}$\Yale$ \\ [7pt]
$\Yale$Yale University \qquad $\UChi$University of Chicago \qquad
$\TCS$TCS Research 
}

\maketitle
\thispagestyle{fancy}
\fancyhead{}
\lhead{%
    \includegraphics[height=1.1cm]{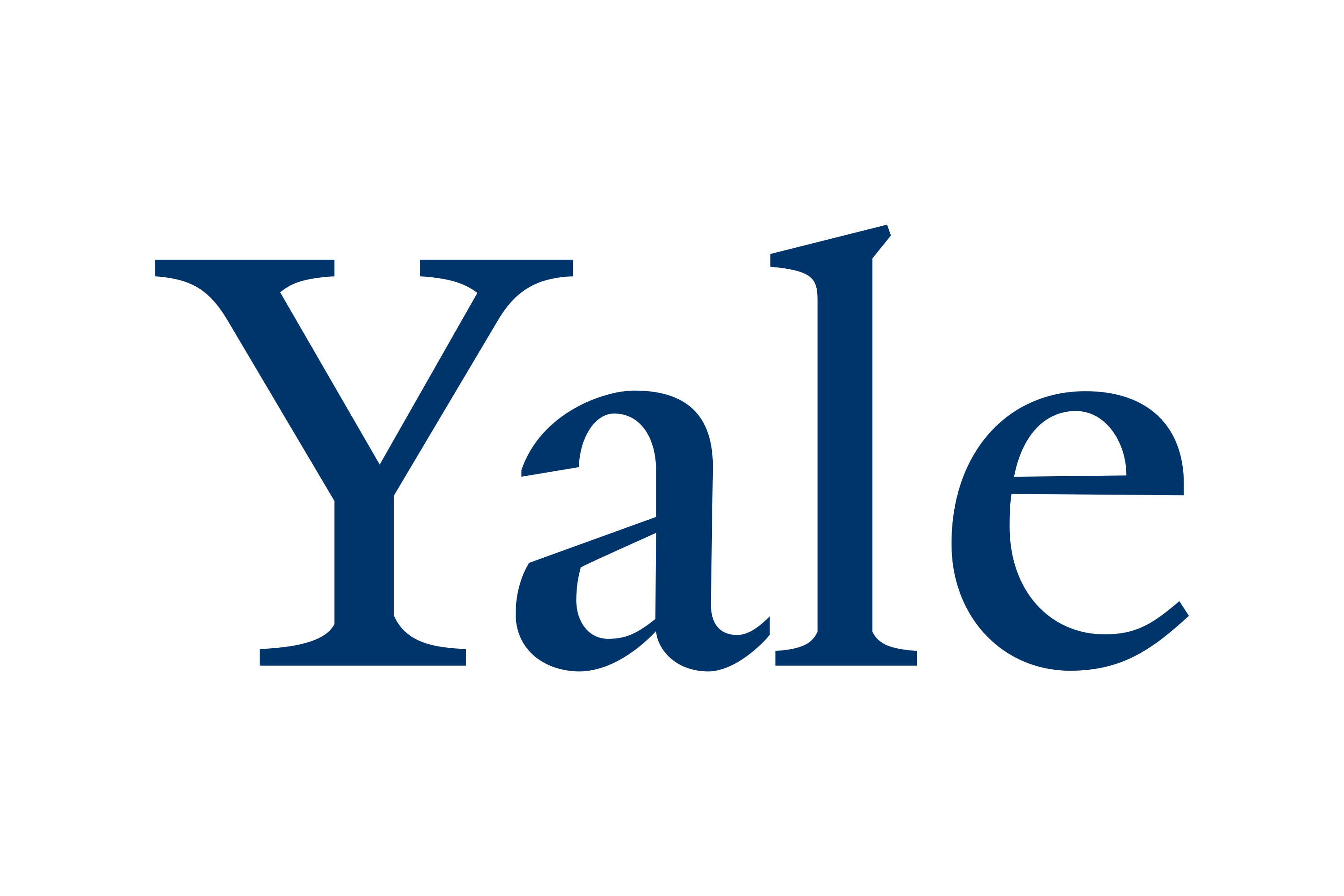}\hspace{0.2cm}%
    \includegraphics[height=1.05cm]{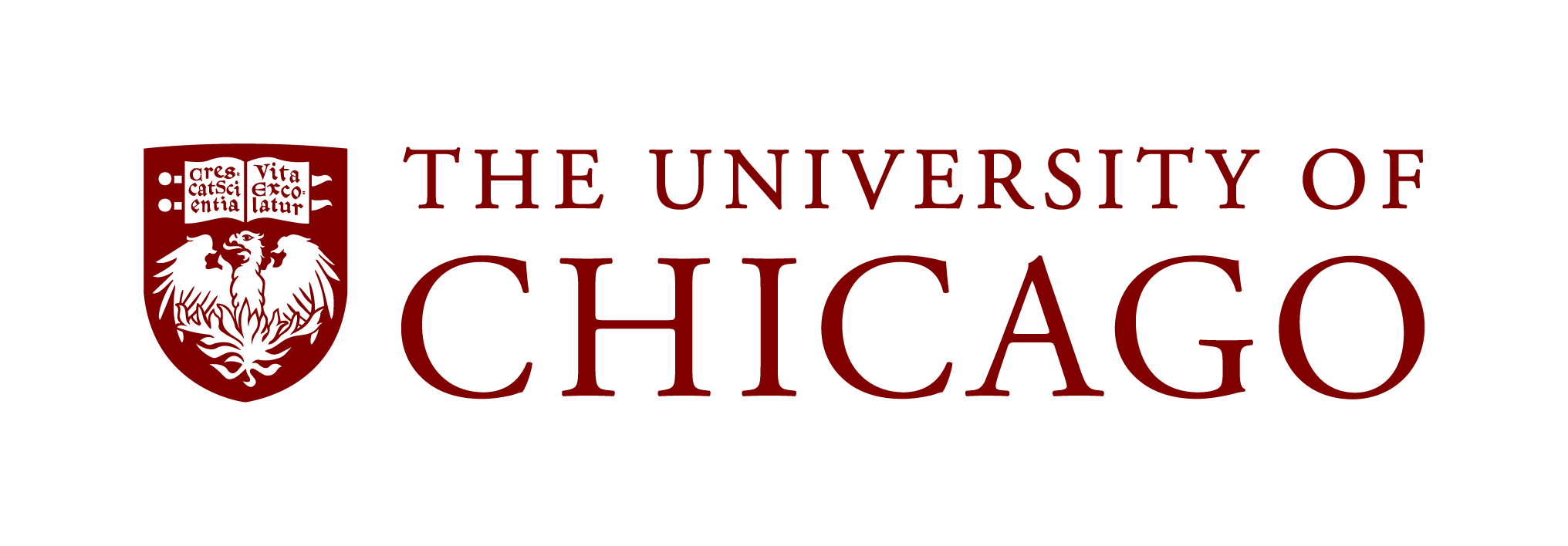}\hspace{0.3cm}%
    \raisebox{0.05cm}{\includegraphics[height=0.8cm]{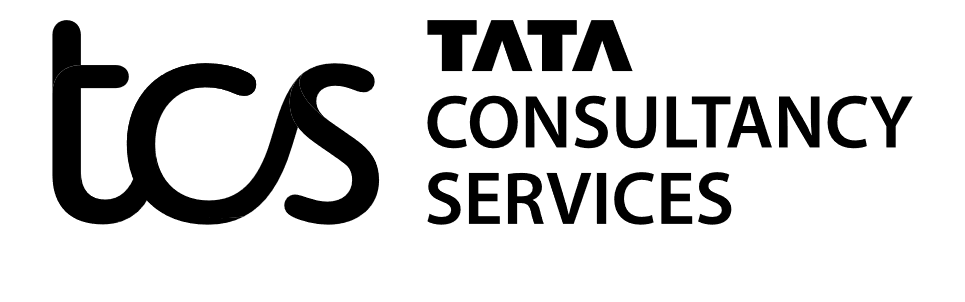}}%
}

\vspace{0.5em}

\fancyfoot[C]{\thepage}
\renewcommand{\headrulewidth}{0pt}
\setlength{\headheight}{12pt}
\addtolength{\topmargin}{0pt}
\setlength{\headsep}{3mm}

\vspace{-0.5em}

\pagestyle{plain}

\begin{abstract}
Constructing scientific multimodal document reasoning datasets for foundation model training  involves an inherent trade-off among scale, faithfulness, and realism. 
To address this challenge, we introduce the \emph{synthesize-and-reground} framework, a two-stage pipeline comprising: (1) \emph{Claim-Centric QA Synthesis}, which generates faithful, isolated QA pairs and reasoning on focused segments, and (2) \emph{Document-Scale Regrounding}, which programmatically re-embeds these pairs into full-document tasks to ensure realistic complexity. 
Using this framework, we construct \data{}, a large-scale training dataset for cross-modal comprehension, comprising 300K QA pairs with explicit reasoning chains across 20K scientific papers.
We further construct \eval{}, an expert-annotated benchmark to evaluate multimodal comprehension within full-length scientific workflows.
Experiments demonstrate that models fine-tuned on \data{} achieve significant improvements across multiple scientific QA benchmarks (\eg ChartQA, SPIQA, \eval{}), particularly in those tasks requiring complex document-level reasoning.

\begin{center}
\begin{tabular}{cl@{\hspace{5em}}cl}
\huggingface & \href{https://huggingface.co/scimdr}{\data} &
\github & \href{https://github.com/ziyuuc/SciMDR}{\data}
\end{tabular}
\end{center}
\vspace{5pt}

\end{abstract}

\begin{figure*}[ht]
\centering
\includegraphics[width=1.0\linewidth]{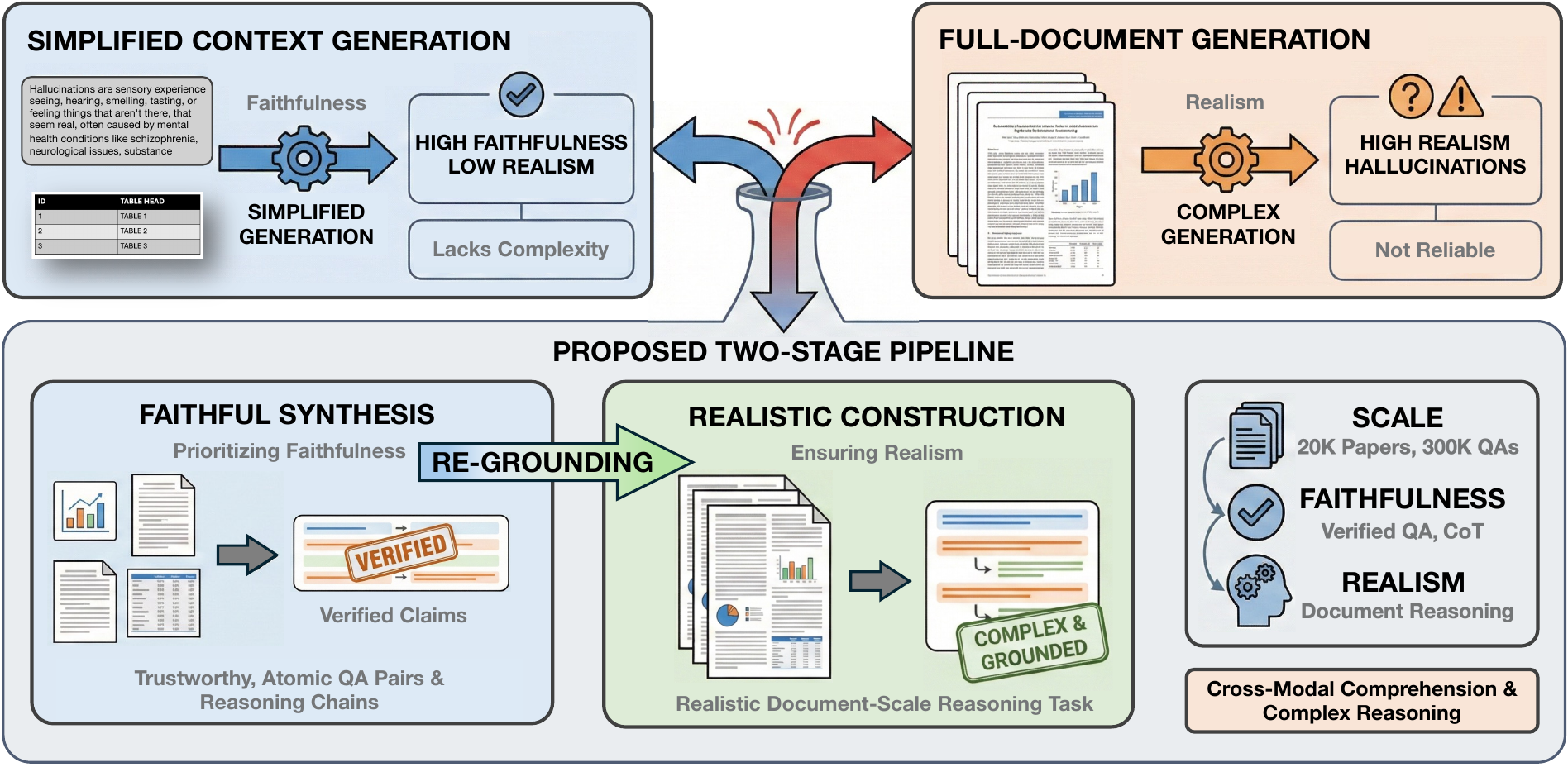}
\caption{\textbf{The Faithfulness-Realism Dilemma in scientific data synthesis and our proposed solution.} Existing approaches face an inherent trade-off: simplifying context ensures \emph{faithfulness} but lacks real-world complexity, while generating directly from full documents ensures \emph{realism} but risks hallucination. We resolve this by decoupling the objectives into a two-stage \emph{synthesize-and-reground} framework. By first generating verified QA pairs on atomic contexts and subsequently re-embedding them into full-document tasks, we achieve a dataset that simultaneously satisfies \emph{Scale, Faithfulness, and Realism}.}
\label{fig:intro}
\end{figure*}

\section{Introduction}
While rapid publication accelerates the spread of ideas, it also makes it harder to locate the most consequential results and to integrate them into coherent understanding~\cite{bornmann2015growth, kusumegi2025scientific}.
LLMs and their multimodal counterparts (\ie MLLMs) offer a promising way to navigate this flood of information, providing tools to quickly summarize synthesize, and query scientific knowledge~\cite{taylor2022galactica, luo2025llm4sr}. 
However, scientific papers remain difficult for general-purpose models because evidence is distributed across long, multimodal documents (text, figures, and tables) and often requires domain expertise to interpret specialized terminology and connect claims to supporting context~\cite{song2025evaluating, wang-etal-2025-sciver, zhao-etal-2025-abgen}. As a result, current models still struggle to provide reliable assistance in real scientific workflows~\cite{zhao2025sciarena, tang2025cellforge, xu-etal-2025-llms-identify}.

A primary reason for this limitation is a deficit in high-quality training data that mirrors the complexity of real-world scientific inquiry. 
This data gap is reflected in the existing Scientific QA (SciQA) datasets. Early efforts rely on costly human annotation
and remained small-scale and often text-only ~\cite{dasigi2021dataset, malaviya2024expertqa,wadden-sciriff}. 
Subsequent work turned to visual elements but adopted a \textit{sanitized-context} approach, focusing on isolated figures or tables~\cite{masry2022chartqa, kahou2017figureqa}. 
Recent work have begun to incorporate full-document contexts, presenting models with more realistic, \textit{in-the-wild} tasks~\cite{pramanick2024spiqa}. This shift, however, has exposed a deeper, unresolved methodological challenge: a fundamental trade-off between \emph{faithfulness} and \emph{realism} in synthetic data.
Specifically, to achieve \emph{faithfulness}, QA generators can be prompted with concise, atomic contexts, which simplifies the task to yield verifiable outputs. 
However, this setup sacrifices realism as it leaves the generation pipeline underexposed to the full-length, complex documents.
Conversely, to achieve \emph{realism}, querying with lengthy, unprocessed documents can more closely mirrors practical use cases. However, this long-context approach leads to attention dilution, increasing the likelihood of hallucinations~\cite{ji2023survey} and undermining faithfulness in the generated ground-truth answers~\cite{liu2024lost, bai2024longbench}. 

As illustrated in \autoref{fig:intro}, to resolve this \emph{faithfulness-realism dilemma}, we propose a new data synthesis paradigm that decouples faithfulness and realism across two stages. The first stage deliberately reduces data synthesis difficulty by structuring synthesis around isolated, claim-centric units and a backward construction to ensure \emph{faithfulness}, while the second stage reintroduces full-document complexity during training instance construction to achieve \emph{realism}.
Specifically, our approach first prioritizes \textit{faithfulness} through synthesis stage. By operating on small, verifiable, and atomic contexts, this stage allows a generator to reliably produce grounded QA pairs and their detailed Chain-of-Thought (CoT) rationales~\cite{wei2022chain}. By constraining the core task and minimizing auxiliary demands, the generator is better positioned to produce trustworthy outputs.
Second, we address \textit{realism} via a training instance construction stage. We re-embed this golden QA-CoT pair within its original, full-document context. 
This design is the key to our solution: the model is presented with a realistic, \textit{in-the-wild} task, but is simultaneously equipped with the precise CoT as ground truth. This demonstration teaches the model both \textit{how to find the evidence} and \textit{how to use evidence to answer questions}, bridging the gap between faithful synthesis and realistic application.
Using this pipeline, we construct \data{}, a new, large-scale (300K QA pairs from 20K papers) dataset for multimodal scientific document reasoning, enabling models to be trained to help users understand central claims, supporting evidence, mechanisms, and comparisons under realistic full-document conditions.

To comprehensively evaluate model performance in real-world scientific scenarios, we construct \eval, a benchmark comprising 907 human annotated QA pairs requiring evidence localization within lengthy, noisy documents, which further enables us to investigate the impact of long-context noise on model robustness.
To validate our approach, we fine-tune \texttt{Qwen2.5-VL-7B} and \texttt{LLaVA-1.5-7B} on \data{}. 
Our empirical evaluation shows that this model significantly outperforms baselines across a comprehensive suite of three established benchmarks (\ie ChartQA~\cite{masry2022chartqa}, CharXiv~\cite{wang2024charxiv} and SPIQA~\cite{pramanick2024spiqa}) and \eval. 
Ablation studies confirm the value of the high-quality reasoning chains within our generated data, and experimental results validate that such data effectively teaches models the skills required for real-world scientific QA. Our main contributions are summarized below: 
\begin{itemize}
\item We introduce \textbf{\eval{}}, an expert-annotated benchmark designed to evaluate model performance in realistic, in-the-wild scientific QA scenarios (\S\ref{sec:bench}).

\item We propose a novel \emph{synthesize-and-reground} paradigm that resolves \emph{faithfulness-realism dilemma} in synthetic data generation by decoupling data generation from training instance construction, ensuring both atomic precision and holistic realism (\S\ref{sec:method}). 

\item We release
\textbf{\data{}}, a large-scale training dataset, by using the designed data synthesis pipeline (\S\ref{sec:method}). 

\item Experiments show that fine-tuning on \data{} improves scientific QA performance, and analyses further confirm that our data provides strong training signals for robust, in-the-wild multimodal reasoning under long-context noise (\S\ref{sec:exp}).

\end{itemize}

\section{Related Work}
\begin{table*}[!t]
  \small
  \centering
  \renewcommand{\arraystretch}{1.0} 
  \caption{\textbf{Comparison of Scientific QA Benchmarks \& Datasets.} Unlike prior works that rely on sanitized contexts or lack reasoning annotations, \data{} integrates \emph{Full-Text} understanding, \emph{Visual} modality, and explicit \emph{chain-of-thought} reasoning at \emph{scale}, bridging the gap between faithful synthesis and realistic document complexity.}
  \resizebox{1.0\linewidth}{!}{
  \begin{tabular}{@{}c|l c c r r c c c@{}}
    \toprule
    \bfseries Category & \bfseries Data &
    \bfseries \begin{tabular}[c]{@{}l@{}}CoT\end{tabular} &
    \bfseries \begin{tabular}[c]{@{}l@{}}Q-Gen\end{tabular} &
    \bfseries Num QA &
    \bfseries \begin{tabular}[c]{@{}c@{}}Source\end{tabular} &
    \bfseries Domain &
    \bfseries \begin{tabular}[c]{@{}c@{}}Full Text\end{tabular} &
    \bfseries \begin{tabular}[c]{@{}c@{}}Visual\end{tabular} \\
    \midrule

    \multirow{8}{*}{\bfseries Bench.} 
    & QASPER~\cite{dasigi2021dataset}       &  & \texttt{human}     & 5K    & 1.5K papers  & NLP   & \xmark & \xmark \\
    & QASA~\cite{lee2023qasa}         &  & \texttt{human}     & 1.8K  & 112 papers   & AI/ML & \cmark & \xmark \\
    & ArgSciChat~\cite{ruggeri2023dataset}   &  & \texttt{human}     & 41    & 20 papers    & NLP   & \cmark & \xmark \\
    & MMLongBench-Doc~\cite{ma2024mmlongbench}      & - & \texttt{human + llms}      & 2.5K   & 1612 charts   & STEM  & \cmark & \cmark \\
    & CharXiv~\cite{wang2024charxiv}      &  & \texttt{human}     & 11.5K & 2.3K charts  & STEM  & \xmark & \cmark \\
    & ChartQAPro~\cite{masry2025chartqapro}   &  & \texttt{human + llms}      & 1.9K    & 1.3K charts  & STEM  & \xmark & \cmark \\
    & DomainCQA~\cite{zhong2025domaincqa}    &  & \texttt{llms}       & 1.7K  & 482 charts   & STEM  & \xmark & \cmark \\

    \cmidrule{2-9} 
    
    & \textbf{\eval}& - & \texttt{human} & 907 & 200 papers & STEM & \cmark & \cmark \\
    
    \midrule
    \midrule

    \multirow{5}{*}{\bfseries DataSet.} 
    & ChartQA~\cite{masry2022chartqa}      & \xmark & \texttt{human + llms}      & 23K   & 28K charts   & STEM  & \xmark & \cmark \\
    & ArXivQA~\cite{li2024multimodal}      & \cmark & \texttt{GPT-4}            & 100K  & 32K charts   & STEM  & \xmark & \cmark \\
    & MMSci~\cite{li2024mmsci}        & \xmark & \texttt{GPT-4}            & 1M     & 128K papers  & STEM  & \xmark & \cmark \\
    & SPIQA~\cite{pramanick2024spiqa}        & \cmark & \texttt{human + llms}      & 270K  & 25.5K papers & CS    & \xmark & \cmark \\
    
    \cmidrule{2-9} 
    
    & \textbf{\data}& \cmark & \texttt{GPT-5.1} & 300K & 20K papers & STEM & \cmark & \cmark \\
    
    \bottomrule
  \end{tabular}
  }
  \label{tab:qa_dataset_comparison}
\end{table*}

Crafting datasets to benchmark and enhance the scientific reasoning capabilities of LLMs necessitates a balance of three critical attributes: \textit{scale}, \textit{faithfulness}, and \textit{realism}. However, achieving this balance presents a fundamental dilemma for prior work. As the general capabilities of LLMs have advanced, their expanding knowledge base offers opportunities for large-scale data synthesis. Yet, existing approaches often compromise one attribute to optimize the others, as summarized in \autoref{tab:qa_dataset_comparison}.

\paragraph{Human-Annotated SciQA.}
Early scientific QA datasets relied on manual annotation to overcome the challenge of generating diverse, open-ended and domain-specific questions.
Initial efforts like PubMedQA~\cite{jin2019pubmedqa}, BioASQ~\cite{krithara2023bioasq}, and QASPER~\cite{dasigi2021dataset} yielded thousands of examples but were often limited to abstracts or fixed formats. Subsequent work, such as QASA~\cite{lee2023qasa} and Covid-QA~\cite{moller2020covid}, utilized full-text annotation for free-form questions, while ExpertQA~\cite{malaviya2024expertqa}, SCIDQA~\cite{singh2024scidqa}, and MISS-QA~\cite{zhao-etal-2025-multimodal-foundation} further enhanced question complexity.
While human annotation typically ensures quality, it faces a bottleneck in \textit{scale}. The expensive nature of expert annotation limits these datasets' size, making them insufficient for training modern foundation models that require vast quantities of data.

\paragraph{Sanitized-Context SciQA.}
With the development of visual capabilities in LLMs, attention has increasingly turned to the visual context within scientific documents, such as figures and tables. Datasets such as DVQA~\cite{kafle2018dvqa}, FigureQA~\cite{kahou2017figureqa}, PlotQA~\cite{methani2020plotqa}, ChartQA~\cite{masry2022chartqa}, and ChartQAPro~\cite{masry2025chartqapro} were proposed to benchmark with QA centered on visual contexts, placing new demands on the models' visual understanding and reasoning.
More recently, MathVista~\cite{lu2023mathvista} and ArXivQA~\cite{li2024multimodal} have further broadened this task's scope by incorporating more charts and diagrams. However, these datasets typically operate on \textit{sanitized contexts}, isolating visual elements from their surrounding textual analysis. This approach creates a discrepancy between the benchmark task and the real-world challenge of navigating noisy, long-form documents. By simplifying the information retrieval process to isolated snippets, these methods compromise \textit{realism}, failing to reflect the complexity of holistic scientific reasoning.

\paragraph{Long-Context SciQA.}
In real-world cases, users frequently query with long, complex documents. Driven by the extension of context windows in LLMs~\cite{team2024gemini, liu2025comprehensive}, many datasets have begun to focus on models' ability to process and answer questions based on long-context. 
For instance, SciREX~\cite{jain2020scirex} is a document-level information extraction dataset, QuALITY~\cite{pang2022quality} involves annotated QA over complete passages, and MMLongBench-Doc~\cite{ma2024mmlongbench} and M3SciQA~\cite{li2024m3sciqa} incorporate visual information and multi-document reasoning through expert curation. The reliance on human annotators constrains the \textit{scale} of these datasets. To address scalability, benchmarks like SPIQA~\cite{pramanick2024spiqa}, Loong~\cite{wang2024leave} and LongReason~\cite{ling2025longreason} typically synthesize questions based on short contexts, introducing extended noise documents during the evaluation. While providing final answers suffices for \textit{benchmarking}, effective \textit{training} demands explicit \textbf{reasoning} that guide models to locate evidence and filter noise. Originating from sanitized contexts, existing synthetic data inherently lacks these global traces, limiting its utility in enhancing \textit{needle-in-a-haystack} reasoning capabilities.

\section{\eval Benchmark}
\label{sec:bench}
We focus on document-level scientific QA, where models must comprehend lengthy, multimodal documents in realistic scenarios. However, existing benchmarks mainly evaluate models on sanitized contexts—isolated figures, tables, or short passages. To bridge this gap and provide an evaluation of models' capabilities in \textit{in-the-wild} scientific reasoning, we construct \eval, an expert-annotated benchmark specifically designed to evaluate document-level multimodal QA performance. This benchmark serves dual purposes: (1) demonstrates the difficulty of in-the-wild scientific reasoning, and (2) provide a general, reliable testbed for evaluating multimodal document understanding in real-world scientific scenarios.

\subsection{Benchmark Construction} \eval is constructed through human annotation to ensure the quality and accuracy. We recruited three annotators (graduate students in computer science) to manually craft QA pairs from 300 scientific papers sourced from arXiv. 
To ensure coverage of scientific reasoning capabilities, we define five question types based on established practices in scientific inquiry and our analysis of real-world SciQA requirements:
\begin{itemize}
    \item \emph{Evidence-Based Explanation \& Quantification}: Explaining \textit{how} and \textit{why} visual element supports textual claim, often with quantitative analysis.
    \item \emph{Concept-to-Instance Mapping}: Linking abstract concepts, architectures, or processes described in text to their concrete visual representations. 
    \item \emph{Hypothesis Validation \& Inferential Reasoning}: Using textual and visual evidence to validate hypotheses, infer conclusions, or predict outcomes.
    \item \emph{Critical Analysis \& Consistency Check}: Critically evaluate whether textual claims are accurately supported by visual data, identifying potential inconsistencies or mischaracterizations.
    \item \emph{Argumentative Role \& Synthesis}: Synthesizing the overall scientific contribution and understanding the role of visual evidence in main argument.
\end{itemize}

For each assigned paper, the annotator was instructed to read the paper and formulate questions that necessitate synthesizing information across both textual content and visual elements distributed throughout the paper. Each entry was authored by one annotator and verified by the other two. 
Annotators were instructed to balance the questions across all types and provided with detailed guidelines and examples to ensure consistency and quality. Annotators also marked key points in each answer to facilitate fine-grained evaluation.
This process yielded 907 high-quality QA pairs with detailed reasoning chains and answer key points for evaluation.

\subsection{Evaluation Protocol} 
Given the open-ended nature of our questions, exact-match and binary score might be inappropriate. Instead, we employ \texttt{GPT-5-mini} as an LLM judge to evaluate model responses. LLM-assisted evaluations are commonly used in many benchmarks~\cite{lu2023mathvista, yu2023mm, wang2024charxiv}. The judge is provided with the question, annotated answer with key points, and response with reasoning chain. It assigns scores based on factual correctness, reasoning quality, and coverage of key points. We provide the implementation details in \autoref{sec:exp_details}.

\begin{table}[t]
    \centering
    \caption{Detailed statistics of the \eval{} benchmark (left) and \data{} training dataset (right). \eval{} is categorized by reasoning type, a taxonomy that also guides the synthesis of multi-modal samples in \data{}; while \data{} is categorized by modality}
    \begin{subtable}[t]{0.48\linewidth}
        \centering
        \setlength{\tabcolsep}{2pt}
        \resizebox{\linewidth}{!}{
        \begin{tabular}{llr}
            \toprule
            \multicolumn{3}{c}{\textbf{Part I: \eval{} (Benchmark)}} \\
            \midrule
            \textbf{Type} & \textbf{Focus} & \textbf{Count} \\
            \midrule
            EEQ & Explanation \& quantitative analysis & 205 \\
            CIM & Linking abstract concepts to visuals & 240 \\
            HVI & Inferential reasoning \& prediction & 244 \\
            CAC & Consistency check \& critical evaluation & 97 \\
            ARS & Synthesis of argument \& visual role & 121 \\
            \midrule
            \textbf{Total} & & \textbf{907} \\
            \bottomrule
        \end{tabular}
        }
        \label{tab:data_statistics_eval}
    \end{subtable}
    \hfill
    \begin{subtable}[t]{0.48\linewidth}
        \centering
        \setlength{\tabcolsep}{2pt}
        \resizebox{\linewidth}{!}{
        \begin{tabular}{llr}
            \toprule
            \multicolumn{3}{c}{\textbf{Part II: \data{} (Training Dataset)}} \\
            \midrule
            \textbf{Category} & \textbf{Description} & \textbf{Count} \\
            \midrule
            TQA & Answerable solely from textual context & 47,389 \\
            VQA & Answerable solely from figures/tables & 125,052 \\
            MQA & Requires synthesis of text and visuals & 132,020 \\
            \midrule
            \textbf{Total} & & \textbf{304,461} \\
            \bottomrule
        \end{tabular}
        }
        \label{tab:data_statistics_data}
    \end{subtable}
    \label{tab:data_statistics}
\end{table}

\section{Training Data Synthesis Pipeline}
\label{sec:method}

\begin{figure*}[!t]
\centering
\includegraphics[width=1.0\linewidth]{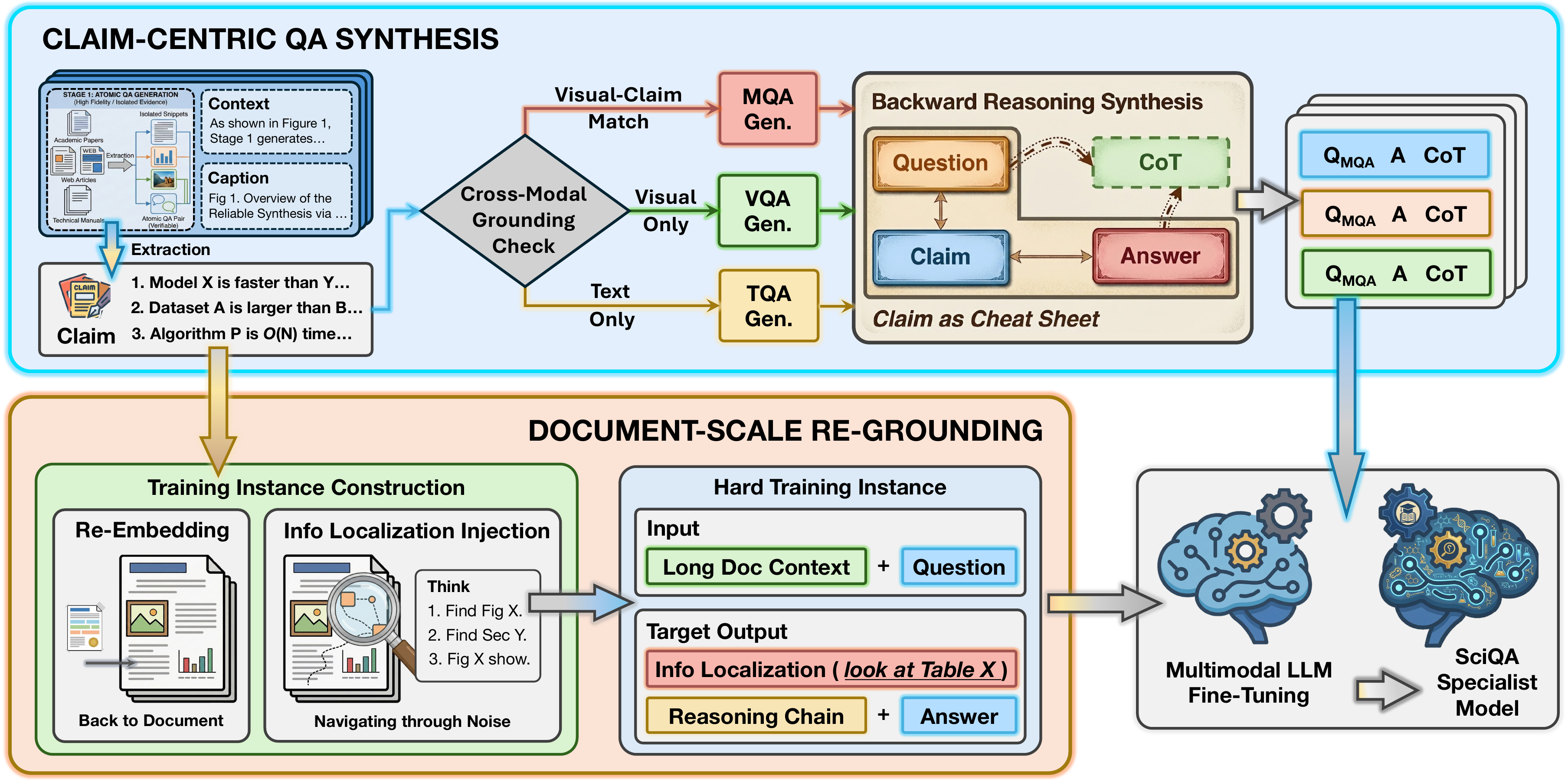}
\caption{\textbf{Overview of the \emph{synthesize-and-reground} framework.} The pipeline operates in two stages: \emph{Claim-Centric QA Synthesis} ensures faithfulness by extracting atomic claims and employing backward reasoning to generate QA pairs with chain-of-thought; \emph{Document-Scale Re-grounding} ensures realism by re-embedding these pairs into full-document contexts and injecting information localization steps to create hard training instances.}
\label{fig:pipeline}
\end{figure*}

To resolve the aforementioned \emph{faithfulness-realism dilemma}, we introduce a two-stage paradigm that decouples data synthesis process from training instance construction, as outlined in \autoref{fig:pipeline}:
\begin{itemize}
\item \textbf{Claim-Centric QA Synthesis:} We first generate high-quality, trustworthy data by reducing the task difficulty for the generator model to ensure correctness and traceability. 
\item \textbf{Document-Scale Regrounding:} Then use this data to construct complex, realistic training instances for full-document comprehension.
\end{itemize} 
This approach allows us to achieve all three goals: generated at scale, high-faithfulness content, and formatted for realistic, complex training.

\subsection{Scientific Paper Collection and Processing}

We collected raw academic papers from two primary sources to construct our dataset: CoRR in arXiv and Nature Communications. Papers from arXiv focus on the Computer Science, comprising a total of 9,847 papers ranging from 2017 to 2025. To ensure our dataset reflects the most recent research advancements, we prioritized papers from the last three years (2023–2025), which constitute over 97\% of our arXiv subset. We also gathered 9,273 General Science articles from Nature Communications published between 2018 and 2025, ensuring a broad coverage of high-quality scientific content.
To parse the multimodal content of each paper, we use the MinerU2.5 OCR model~\cite{niu2025mineru} with a vLLM backend. Given a downloaded PDF, our adopted OCR pipeline extracts the full body text, section boundaries, figures, tables, and associated captions. We serialize these outputs into JSON files, which are then used by the subsequent data synthesis pipeline.
For each paper, we then use \texttt{GPT-5.1} to assess whether it reports an original, experiment-driven study, filtering out surveys, position papers, tutorials, and purely conceptual work. \autoref{tab:data_statistics} presents a detailed breakdown of the resulting dataset statistics.

\subsection{Claim-Centric QA Synthesis}
The objective of this stage is to produce a corpus of trustworthy, atomic QA pairs and their corresponding reasoning chains, all grounded in the source document. We achieve this quality by operating on small and isolated contexts, and employing a claim-centric mechanism. QA pairs can be classified into three types based on the information source required for an answer: \textbf{VQA} (\textit{Vision-Only QA}), answerable solely from visual information (figures and tables); \textbf{TQA} (\textit{Text-Only QA}), answerable solely from textual context; and \textbf{MQA} (\textit{Multi-modal QA}), which requires synthesizing information from both text and visuals. Each category is further defined by specific sub-types to balance generation diversity with controllability.

Our synthesis process begins with a multi-modal context unit, each comprising a segment of raw text, an associated visual (figure or table), and its caption. The core of this process is a claim-centric mechanism. We first perform a context-aware pre-processing step to identify all sentences within the text that reference the associated visual (\eg \textit{As shown in Figure X...}). We then feed the processed text into the LLM generator. At this time, the visual information is temporarily withheld to ensure a purely text-based analysis. Our prompt marks the previously identified referencing sentences, prioritizing segments most likely to contain arguments for later visual grounding. Following this guidance, the LLM generator breaks down the text into discrete, declarative claims, each representing a core finding or conclusion.

These extracted claims then serve as the unified \emph{blueprint} for both QA and reasoning synthesis. First, there is a cross-modal grounding step, the LLM generator revises its claims by checking each one against the previously withheld visual information to determine whether a direct visual correlate exists. Claims with visual correlates are routed for MQA generation, text-only claims are routed for TQA, and VQA pairs are generated in parallel by focusing the LLM exclusively on the visual. Besides, for each QA pair, we guide the generation of its reasoning chain. We reframe this from a inference task to a low-risk, constrained articulation task. The claim is the key to this shift, acting as a cheating sheet with the ground-truth conclusion. By providing this answer upfront, we transform the task of LLM generator from finding an answer, to articulate a step-by-step rationale that logically connects a newly generated question to the supplied claim. This backward construction paradigm makes the synthesis easy by offloading tasks of evidence retrieval and open-ended inference, yielding reasoning chains both trustworthy and controllable.

\subsection{Document-Scale Regrounding}
The atomic QA pairs and reasoning chains generated on small and isolated contexts are suited for benchmarking a model's capabilities, but they are suboptimal as training data. This is because, in realistic application scenarios, users rarely filter relevant paragraphs before posing a query. Instead, the more common use case involves interrogating the entire noisy, complex document. Simply training on the atomic QA pairs would fail to prepare the model for this full-context challenge.

We bridge this gap by re-purposing claims. The claim, which served as a generation blueprint in synthesis, now functions as a ground-truth evidence map for the training stage. 
Because each QA pair is bound to a claim, which records the precise location of its textual and visual evidence, we can programmatically construct an ideal \textit{Information Localization} step. This is achieved by populating a pre-defined template with the specific identifiers (\eg \textit{Section X, Table }Y) stored in the claim. This content, which explicitly states how to find the necessary information, is then prepended to the synthetic reasoning chains. 
For example: \textit{To answer this question, I need to first consult Section X, and then cross-reference the results in Table Y...}

This deterministic synthesis of CoT rationales provides the downstream model with an accurate, verifiable, and imitable reasoning demonstration. This creates the hard training instance: the task is no longer a simple query on a filtered easy context, but a realistic, hard challenge that requires finding evidence within the full document. Critically, while the task difficulty is high, the solutions we provide via demonstrations are detailed and well-structured. With such data, the model is not just learning \textit{what} the answer is; it is learning \textit{how to find the answer} within a complex context. The final training data format is structured as: \textit{(Full Document Context, Question)} $\rightarrow$ \textit{(Information Localization + Reasoning + Final Answer)}. This format compels the model to first practice localizing related information and then execute grounded reasoning, thereby enhancing its practical utility in real-world scientific QA applications.

\section{Experiments}
\label{sec:exp}

We conduct experiments to verify the effectiveness of our proposed data construction pipeline and \data{}, addressing two research questions:

\begin{itemize}
\item \textbf{RQ1:} Does fine-tuning on \data{} enhance model performance on scientific reasoning? 
\item \textbf{RQ2:} Does our synthetic data pipeline possess the capability to produce useful training data that improves model scientific reasoning?
\end{itemize} 

\subsection{Experimental Setup}

\paragraph{Dataset.} Our dataset \data{} comprises three categories based on information sources: VQA, TQA, and MQA. The dataset was constructed following the pipeline in Section~\ref{sec:method}, generating approximately 300K QA pairs with claim-centric reasoning chains from 20K research papers with \texttt{GPT-5.1}.

\paragraph{Training Configuration.} We employ a two-stage training, using \texttt{Qwen2.5-VL-7B}~\cite{bai2025qwen2} as our primary base model.
In Stage 1, we train on VQA and TQA data for 1 epoch with a peak learning rate $1 \times 10^{-5}$ and batch size 64. In Stage 2, we continue training on MQA data for 1 epoch with learning rate $1 \times 10^{-6}$. In fine-tuning with SPIQA, we train the language model for 1 epoch with a learning rate of $1 \times 10^{-5}$ and batch size 64. We fine-tune the language model while keeping the visual encoder and projector frozen.

\paragraph{Evaluation Benchmarks.} We evaluate models on four benchmarks: (1) \textbf{ChartQA}~\cite{masry2022chartqa}, a foundational chart QA benchmark to evaluate logical and visual reasoning over standard real-world charts; (2) \textbf{CharXiv}~\cite{wang2024charxiv}, a benchmark for scientific QA that uses expert-curated charts from research papers to assess both \textbf{\textit{D}}escriptive examination and complex \textbf{\textit{R}}easoning capabilities; (3) \textbf{SPIQA}~\cite{pramanick2024spiqa}, a benchmark with 3 subsets designed to assess multimodal comprehension of academic content, which requires a holistic understanding of complex figures and tables within full-text papers; and (4) \textbf{\eval}, our annotated benchmark for full-document scientific reasoning.

\paragraph{Baselines.} We benchmark our method against the base model \texttt{Qwen2.5-VL-7B} to measure relative gains, and reproduce SPIQA, a recent synthetic baseline, by fine-tuning the same base model to isolate data quality effects. We also include several strong open-source multimodal models \texttt{Qwen-3-VL-8B}~\cite{Bai2025Qwen3VLTR}, \texttt{LLaVA-OV-1.5-8B}~\cite{An2025LLaVAOneVision15FO}, and \texttt{InternVL-3-8B}~\cite{Zhu2025InternVL3EA} as competitive references. In addition, we evaluate some advanced models \texttt{GPT-4o}~\cite{gpt4o}, \texttt{GPT-5.1}~\cite{openai_gpt5.1_2025}, and \texttt{GPT-5.2}~\cite{openai_gpt5.2_2025} on \eval to establish a performance upper bound and analyze the development of scientific multimodal document reasoning capability.

\begin{table*}[!t]
\small
\centering
\caption{\textbf{Main results on scientific QA benchmarks.} Fine-tuning with \data{} outperforms the base model and the recent synthetic dataset across most metrics, particularly on complex reasoning tasks.}
\resizebox{1.0\linewidth}{!}{
\begin{tabular}{lccccccc}
\toprule
\textbf{Model} & \textbf{ChartQA} & \textbf{CharXiv-D} & \textbf{CharXiv-R} & \textbf{SPIQA-A} & \textbf{SPIQA-B} & \textbf{SPIQA-C} & \textbf{\eval} \\
\midrule
\texttt{GPT-5.1} & - & 90.9 & 58.3 & 79.4 & 79.8 & 71.6 & 47.2 \\
\texttt{GPT-5.2} & - & 95.2 & 73.1 & 79.9 & 75.4 & 74.0 & 49.9 \\
\texttt{Qwen-3-VL-8B} & 87.4 & 74.2 & 40.1 & 73.2 & 64.0 & 62.3 & 34.2 \\
\texttt{LLaVA-OV-1.5-8B} & 85.9 & 66.3 & 32.9 & 66.0 & 62.7 & 51.1 & 15.5 \\
\texttt{InternVL-3-8B} & 86.2 & 66.7 & 34.6 & 59.6 & 46.9 & 40.8 & 16.8 \\
\midrule
\texttt{Qwen2.5-VL-7B} & 84.6 & 65.0 & 37.7 & 66.4 & 56.6 & 48.9 & 19.8 \\
\midrule
\quad + SPIQA 
& 81.8\rlap{\deltaScore{-2.8}}
& 50.9\rlap{\deltaScore{-14.1}}
& 33.3\rlap{\deltaScore{-4.4}}
& 62.7\rlap{\deltaScore{-3.7}}
& 44.7\rlap{\deltaScore{-11.9}}
& 40.0\rlap{\deltaScore{-8.9}}
& 5.6\rlap{\deltaScore{-14.2}} \\

\quad + \textbf{\data{}} 
& 86.3\rlap{\deltaScore{+1.7}} 
& 75.6\rlap{\deltaScore{+10.6}} 
& 37.9\rlap{\deltaScore{+0.2}} 
& 68.6\rlap{\deltaScore{+2.2}} 
& 58.8\rlap{\deltaScore{+2.2}} 
& 47.3\rlap{\deltaScore{-1.6}} 
& 49.1\rlap{\deltaScore{+29.3}} \\
\bottomrule
\end{tabular}
}
\label{tab:main_results}
\end{table*}

\subsection{Main Results}

\begin{wraptable}{r}{0.40\textwidth}
\vspace{-1.1\baselineskip}
\centering
\caption{\textbf{Performance comparison on \eval against advanced models.} Despite having only 7B parameters, our model matches the performance of \texttt{GPT-5.2} and \texttt{GPT-5.1} on this domain-specific task.}
\small
\begin{tabular}{lc}
\toprule
\textbf{Model} & \textbf{\eval} \\
\midrule
\texttt{GPT-5.2} & \textbf{49.9} \\
\texttt{GPT-5.1} & 47.2 \\
\texttt{GPT-4o} & 24.7 \\
\midrule
\texttt{Qwen2.5-VL-7B} & 19.8 \\
\midrule
\quad + \textbf{\data{}} & 49.1\rlap{\deltaScore{+29.3}} \\
\bottomrule
\end{tabular}
\label{tab:sota_comparison}
\vspace{-1.0\baselineskip}
\end{wraptable}

\autoref{tab:main_results} and \autoref{tab:sota_comparison} presents the comparative performance of model fine-tuned with \data, against the baselines across all four benchmarks.
The results substantiate the efficacy of our approach (RQ1). \emph{Model fine-tuned with \data{} achieves substantial improvements over the base model across the board}, effectively transforming a general-purpose multimodal model into a specialized scientific assistant. 
To further contextualize the difficulty of our proposed benchmark and the effectiveness of our method, we compare our fine-tuned model against advanced proprietary models on \eval. Despite its smaller parameter size 7B, model with \data{} exhibits competitive performance on this scientific reasoning task.

\subsection{Pipeline Effectiveness and Analysis}
Having established the performance gains, we address RQ2 by analyzing the quality of our synthetic data and deconstructing the contributions of our pipeline components.

\subsubsection{Data Quality Comparison}
To assess the quality of our synthetic data independent of the base model's intrinsic capabilities, we conduct a controlled comparison using \texttt{LLaVA-1.5-7B}~\cite{liu2024improved}. 
We chose \texttt{LLaVA-1.5} as our probing model for two strategic reasons: its fully transparent training data ensures no prior exposure to our evaluation benchmarks, and as a more modest baseline, it is more sensitive to data quality, allowing us to clearly observe the marginal gains from different instruction-tuning datasets.
We fine-tune \texttt{LLaVA-1.5-7B} on three configurations: (1) 50K samples from SPIQA, (2) 50K VQA samples from \data{}, and (3) 50K samples from SPIQA re-annotated using our claim-centric pipeline. All models are trained for 2 epochs and evaluated on on single-image benchmarks to match the model's input constraint.

\autoref{tab:data_quality} confirms that re-annotating SPIQA with our pipeline outperforms the original labels (39.8 vs. 35.7) using identical source documents. This isolates the gains to our methodology rather than data selection. We attribute this improvement to the rich reasoning signals in our data: notably, the model trained on our re-annotated SPIQA generates responses on CharXiv that are $5\times$ longer on average than the original data, reflecting a substantial enhancement in reasoning depth and details.

\begin{table}[H]
\small
\centering

\begin{minipage}[t]{0.56\linewidth}
\centering
\captionsetup{width=0.95\linewidth}
\caption{\textbf{Controlled data quality comparison.} Re-annotating SPIQA with our pipeline improves performance, demonstrating superior data quality.}
\vspace{0.25em}
\setlength{\tabcolsep}{3pt}
\begin{tabular}{lccc}
\toprule
\textbf{Method} & \textbf{ChartQA} & \textbf{CharXiv} & \textbf{SPIQA-A} \\
\midrule
\texttt{LLaVA-1.5-7B} & 19.6 & 27.8 & 31.5 \\
\midrule
\quad + SPIQA (50k)
& 26.3\rlap{\deltaScore{+6.7}} 
& 13.5\rlap{\deltaScore{-14.3}} 
& 35.7\rlap{\deltaScore{+4.2}} \\
\quad + \textbf{\data{}} (50k) 
& \textbf{26.8}\rlap{\deltaScore{+7.2}} 
& \textbf{28.5}\rlap{\deltaScore{+0.7}} 
& 36.7\rlap{\deltaScore{+5.2}} \\
\quad + SPIQA (re-annotated) 
& 25.5\rlap{\deltaScore{+5.9}} 
& 28.1\rlap{\deltaScore{+0.3}} 
& \textbf{39.8}\rlap{\deltaScore{+8.3}} \\
\bottomrule
\end{tabular}
\label{tab:data_quality}
\end{minipage}
\hfill
\begin{minipage}[t]{0.40\linewidth}
\centering
\captionsetup{width=0.95\linewidth}
\caption{\textbf{Ablation study of training data components.} Both explicit information localization and step-by-step reasoning are critical for successful fine-tuning.}
\setlength{\tabcolsep}{4pt}
\begin{tabular}{ccc}
\toprule
\textbf{Info Loc} & \textbf{Reasoning} & \textbf{\eval} \\
\midrule
\cmark & \cmark & \textbf{49.1} \\
\midrule
\xmark & \cmark & 22.8\rlap{\deltaScore{-26.3}} \\
\xmark & \xmark & 16.9\rlap{\deltaScore{-32.2}} \\
\bottomrule
\end{tabular}
\label{tab:ablation}
\end{minipage}

\end{table}

\subsubsection{Ablation Study on Reasoning Chains}
We further investigate which components of our training data contribute to full-document comprehension. Using the Stage 1 checkpoint, we evaluate three variants on \eval: (1) full data with explicit information localization and reasoning chains, (2) removing localization, and (3) removing reasoning chains (QA pairs only).
\autoref{tab:ablation} reveals that removing reasoning chains leads to a significant drop in performance (49.1 $\rightarrow$ 16.9), underscoring that simple QA pairs are insufficient for teaching complex scientific logic. Removing information localization also causes a drop, indicating that explicit guidance on \textit{where} to look is important for helping models navigate the noise in full-text documents.

\subsubsection{Impact of Long-Context Noise}
\label{sec:exp_noise}

\begin{wraptable}{r}{0.35\textwidth}
\centering
\caption{\textbf{Challenge of Attention Dilution.} Effect of context noise on accuracy. Performance degrades as the amount of irrelevant context increases.}
\small
\setlength{\tabcolsep}{4pt}
\begin{tabular}{lc}
\toprule
\textbf{Input} & \textbf{\eval} \\
\midrule
Standard & 19.8 \\
\midrule
Oracle & 32.9 \\
Full-Paper & 12.8 \\
\bottomrule
\end{tabular}
\label{tab:noise_impact}
\end{wraptable}

Our pipeline is motivated by the observation that generating data directly from long, noisy contexts reduces faithfulness. To empirically quantify the impact of noise, we evaluate \texttt{Qwen2.5-VL-7B} on \eval under three input settings: (1) \textbf{Oracle Context}, which provides only the ground-truth visual and referencing text with zero distractors; (2) \textbf{Standard Setting}, \eval default which simulates realistic retrieval by including limited noise (maximum 8 images and 6 paragraphs); and (3) \textbf{Full-Paper}, which supplies the entire document content to maximize distractor density. \autoref{tab:noise_impact} reveals a clear performance degradation as noise increases. The gap between \textbf{Oracle Context} (32.9) and \textbf{Full-Paper} (12.8) confirms that long-context distractors are a source of error; even when the information is present, the model struggles to localize evidence within dense content.

\subsubsection{Failure Analysis}
\label{sec:failure}

We conduct a failure analysis on \eval{}, comparing predictions from the base model and its fine-tuned counterpart on \data{}. We observe and define four main error types: (1) \emph{incorrect evidence localization}; (2) \emph{reasoning or logical errors}; (3) \emph{hallucination of unsupported context}; and (4) \emph{incomplete synthesis of key points}.

Overall, the fine-tuned model shows clear improvements in grounding and evidence localization, suggesting that the structured reasoning signals in \data{} effectively can reduce hallucination and improve document-level reasoning. Details can be found in \autoref{sec:exp_details}.

\section{Conclusion and Discussion}
In this work, we addressed the \emph{faithfulness-realism dilemma} in constructing synthetic datasets for multimodal scientific document reasoning. 
We introduced the \emph{synthesize-and-reground} framework, which decouples atomic reasoning synthesis from full-document training. With \data{} and \eval{}, we demonstrate that our approach enables open-source models to bridge the performance gap with proprietary systems in complex multimodal document reasoning.
Given reliance on proprietary models and STEM focus, future work will explore distilling synthesis into open-source models and expanding domains.

\section*{Acknowledgements}
This work was supported in part by Google's Research Scholar Program.

\section*{Limitations}
While our \emph{synthesize-and-reground} framework effectively enhances scientific multimodal reasoning, several limitations remain.
The fidelity of our training data is intrinsically bounded by the capabilities of the proprietary teacher model (\texttt{GPT-5.1}) used for atomic synthesis. We assume that breaking the task into atomic claims minimizes hallucinations, yet any subtle factual errors or reasoning flaws generated at this stage are hard-coded into the training signal. In practice, if the teacher model exhibits specific biases or misconceptions regarding niche scientific domains, these will inevitably propagate to the student model.
Regarding the scope of our claims, our empirical validation is concentrated within STEM disciplines (primarily Computer Science and General Science). This focus partly reflects the current scarcity of data resources outside of the hard sciences. Consequently, our results have not yet been validated in fields with distinct reasoning paradigms, such as the Humanities or Social Sciences, where scientific discourse may follow different structures.

\bibliographystyle{unsrtnat}
\bibliography{main}

\newpage
\appendix
\appendix

\section{Data and Experimental Details}
\label{sec:exp_details}

\subsection{Configuration}

\paragraph{\texttt{Qwen2.5-VL-7B}.}

We fine-tuned \texttt{Qwen2.5-VL-7B} using LLaMA-Factory~\cite{zheng2024llamafactory} with the following configurations. The maximum sequence length was set to 16K tokens (including both visual and language tokens) to accommodate long-context scientific documents. For image inputs, we set a maximum of 8 images per instance with \texttt{max\_pixels} = $512 \times 512$. Images are automatically resized to maintain their aspect ratio within the specified pixel range.

For \emph{VQA + TQA}. We trained on visual-only and text-only QA pairs for 1 epoch with learning rate $1 \times 10^{-5}$ and batch size 64. Only the language model was trained while the visual encoder and projector remained frozen.

For \emph{MQA}. We continued training on multimodal QA pairs for 1 epoch with learning rate $1 \times 10^{-6}$ and batch size 64, maintaining the same freeze strategy.

\paragraph{\texttt{LLaVA-1.5-7B}.}

For data quality comparison experiments, we fine-tuned \texttt{LLaVA-1.5-7B} using LLaMA-Factory for 2 epochs with learning rate $1 \times 10^{-5}$, batch size 64, and warmup ratio 0.1. Unlike the Qwen experiments, all model components (vision encoder, projector, and language model) were trained without freezing.

\subsection{Evaluation Framework}
\label{sec:eval_framework}

All evaluations were conducted using \texttt{lmms-eval}~\cite{zhang2024lmmsevalrealitycheckevaluation}, which provides standardized evaluation protocols for large multimodal models. We implemented a custom evaluation module for \eval{} to ensure consistency with existing benchmarks.

\subsection{Human Evaluation of Synthetic Data}
\label{sec:human_eval_syn_data}

To directly assess the faithfulness and reliability of the synthesized training data, we conducted a manual evaluation on a random sample of 300 QA pairs from \data{}. The sample consists of 100 instances from each category: VQA, TQA, and MQA. The sampled QA pairs were manually reviewed by the authors using two criteria: \emph{correctness}, which measures whether the answer is factually accurate, and \emph{relevance}, which measures whether the question-answer pair is properly grounded in the source document and associated visual evidence.

For VQA and TQA, all reviewed instances were factually correct and well grounded in the provided visual or textual context. For MQA, 91 out of the 100 sampled instances were judged to be high-quality and fully accurate. The remaining 9 instances were still factually correct, but were occasionally more verbose or focused on high-level concepts, resulting in longer and more complex reasoning chains.

\subsection{LLM Judge Configuration}
\label{sec:llm_judge}

Given the open-ended nature of questions in \eval{}, we employed \texttt{GPT-5-mini} as an LLM judge to evaluate model responses. The judge assesses each response based on factual correctness, reasoning quality, and coverage of annotated key points.

\paragraph{Binary Scoring.} For main results (\autoref{tab:main_results}), we use strict binary scoring: a response receives score 1 only if it correctly addresses all key points with accurate reasoning; otherwise it receives 0. The accuracy is computed as the percentage of fully correct responses.

\paragraph{Fine-grained Metrics.} For detailed analysis, we also report text correctness rate (percentage correctly interpreting textual evidence), visual correctness rate (percentage correctly interpreting visual evidence), and partial credit score (average proportion of key points addressed). These fine-grained metrics provide additional insight but are not used for main benchmark comparison. The complete judge prompt is provided in \autoref{fig:llm_judge_prompt}.

\subsection{Failure Mode Analysis}
\label{sec:appendix_failure}

\paragraph{Setup.}
To analyze failure patterns, we randomly sampled 100 questions from \eval{} and compared outputs from the base model \texttt{Qwen2.5-VL-7B} and the fine-tuned on \data{}. We manually categorized incorrect predictions into four error types.

\paragraph{Failure Categories.} We define the following error categories:

\begin{itemize}
\item \emph{Incorrect Evidence Localization}: Selecting the wrong visual element or paragraph instead of the true supporting context.
\item \emph{Reasoning / Logic Error}: Correctly locating relevant evidence but failing in multi-step deduction or computation.
\item \emph{Hallucination of Context}: Fabricating numbers, visual features, or statements not present in the document.
\item \emph{Incomplete Synthesis}: Identifying correct evidence but missing key annotated answer points.
\end{itemize}

\paragraph{Findings.}
Both quantitative error analysis and qualitative inspection demonstrate that the structured reasoning signals in \data{} are important for improving multimodal document-level scientific QA. The fine-tuned model benefits from explicit localization supervision and exhibits stronger grounding behavior compared to the base model.

\begin{table}[h]
\small
\centering
\caption{Failure type comparison on 100 randomly sampled \eval{} questions.}
\begin{tabular}{lcc}
\toprule
\textbf{Failure Type} & \textbf{Qwen} & \textbf{\data{}} \\
\midrule
Incorrect Evidence Localization & 18 & 5 \\
Reasoning / Logic Error & 6 & 9 \\
Hallucination of Context & 11 & 3 \\
Incomplete Synthesis & 8 & 7 \\
\midrule
Total Errors & 43 & 24 \\
\bottomrule
\end{tabular}
\label{tab:failure_modes}
\end{table}

\section{Annotator and Data Usage}
\label{sec:annotator}

\paragraph{Annotator Recruitment.} For constructing \eval{}, we recruited three graduate students in Computer Science with at least one year of experience in machine learning research and scientific paper analysis. Annotators were compensated above local minimum wage, consistent with standard research assistant rates. All annotators provided written informed consent before participating.

\paragraph{Consent and Usage Rights.} Prior to annotation, participants received detailed consent forms explaining the research purpose, public data release, withdrawal rights, confidentiality measures, and compensation structure. 

For source papers in \data{} and \eval{}, we exclusively used open-access publications from arXiv (various Creative Commons licenses) and Nature Communications (CC-BY license). These licenses permit text and data mining for research purposes, requiring no additional consent from paper authors.

\paragraph{Quality Control.} To ensure annotation quality, annotators underwent training with detailed guidelines and examples. Each QA pair was authored by one annotator and verified by the other two. Weekly meetings addressed challenging cases and maintained consistency. 

\paragraph{Annotation Cost.} For pre-annotation, the total setup time for the annotation team was approximately 5 hours, which included designing guidelines, creating samples, and conducting a training session to align the annotators with the protocol. For annotation, the average time for reading and annotating a single paper was approximately 10 minutes.

\section{Data Synthesis Prompts}
\label{sec:prompts}

This section presents the complete prompts used in our data synthesis pipeline, corresponding to the stages described in Section~\ref{sec:method}.

\paragraph{Claim Extraction.} \autoref{fig:prompt_claim} shows the prompt that guides the LLM to distill paragraphs into structured, verifiable claims serving as blueprints for QA generation.

\paragraph{Visual Grounding.} \autoref{fig:prompt_vground} presents the prompt for matching textual claims with visual evidence and determining their relationship types.

\paragraph{Multimodal QA Generation.} \autoref{fig:prompt_mqa} details the prompt for generating questions requiring synthesis of textual and visual information across five reasoning types (EEQ, CIM, HVI, CAC, ARS).

\paragraph{Visual-Only QA Generation.} \autoref{fig:prompt_vqa} provides the prompt for generating questions answerable solely from visual information across eight reasoning categories.

\paragraph{Text-Only QA Generation.} \autoref{fig:prompt_tqa} shows the prompt for generating questions testing deep understanding of scientific content without visual evidence.

\begin{figure*}[t]
\small
\begin{tcolorbox}[colback=gray!5, colframe=gray!75!black, boxrule=0.5pt]
\input{section/prompt/tqa}
\end{tcolorbox}
\caption{\textbf{TQA generation prompt.} This prompt generates questions testing deep understanding of scientific content without visual evidence.}
\label{fig:prompt_tqa}
\end{figure*}

\begin{figure*}[t]
\small
\begin{tcolorbox}[colback=gray!5, colframe=gray!75!black, boxrule=0.5pt]
\input{section/prompt/judge}
\end{tcolorbox}
\caption{\textbf{LLM judge prompt.} This prompt evaluates model responses based on text citation (0.30), image citation (0.30), and answer accuracy (0.40).}
\label{fig:llm_judge_prompt}
\end{figure*}

\begin{figure*}[t]
\small
\begin{tcolorbox}[colback=gray!5, colframe=gray!75!black, boxrule=0.5pt]
\input{section/prompt/claim}
\end{tcolorbox}
\caption{\textbf{Claim extraction prompt.} This prompt guides the LLM to distill paragraphs into structured, verifiable claims serving as blueprints for QA generation.}
\label{fig:prompt_claim}
\end{figure*}

\begin{figure*}[t]
\small
\begin{tcolorbox}[colback=gray!5, colframe=gray!75!black, boxrule=0.5pt]
\input{section/prompt/vground}
\end{tcolorbox}
\caption{\textbf{Visual grounding prompt.} This prompt matches textual claims with visual evidence, determining relationship types (Supports, Quantifies, Illustrates, Elaborates, Contradicts).}
\label{fig:prompt_vground}
\end{figure*}

\begin{figure*}[t]
\small
\begin{tcolorbox}[colback=gray!5, colframe=gray!75!black, boxrule=0.5pt]
\input{section/prompt/mqa}
\end{tcolorbox}
\caption{\textbf{MQA generation prompt.} This prompt generates questions requiring synthesis of textual and visual information across five reasoning types (EEQ, CIM, HVI, CAC, ARS).}
\label{fig:prompt_mqa}
\end{figure*}

\begin{figure*}[t]
\small
\begin{tcolorbox}[colback=gray!5, colframe=gray!75!black, boxrule=0.5pt]
\input{section/prompt/vqa}
\end{tcolorbox}
\caption{\textbf{VQA generation prompt.} This prompt generates questions answerable solely from visual information across eight reasoning categories.}
\label{fig:prompt_vqa}
\end{figure*}

\section{MQA Examples}
\label{sec:mqa_examples}

This section presents examples of multimodal QA pairs across the five question types.

\begin{figure*}[t]
\centering
\includegraphics[width=0.95\linewidth]{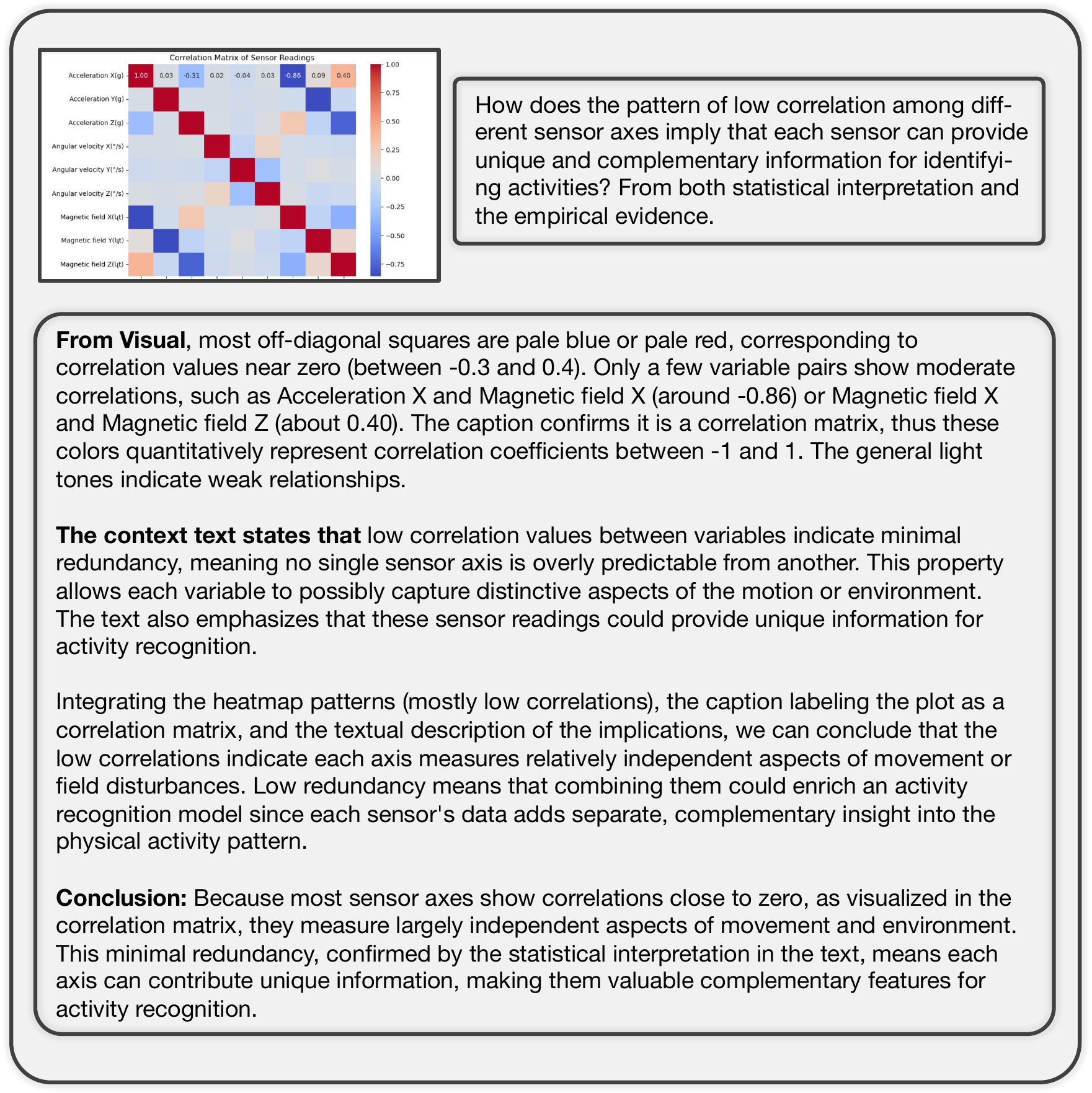}
\caption{\textbf{Example of EEQ (Evidence-Based Explanation \& Quantification) type question.} This example demonstrates how the model must explain how visual patterns (correlation matrix) support textual claims with quantitative analysis, integrating statistical interpretation from the figure with conceptual explanations from the text.}
\label{fig:example_eeq}
\end{figure*}

\paragraph{Evidence-Based Explanation \& Quantification (EEQ).} \autoref{fig:example_eeq} illustrates an EEQ-type question requiring quantitative analysis of visual evidence to support textual claims.

\begin{figure*}[t]
\centering
\includegraphics[width=0.95\linewidth]{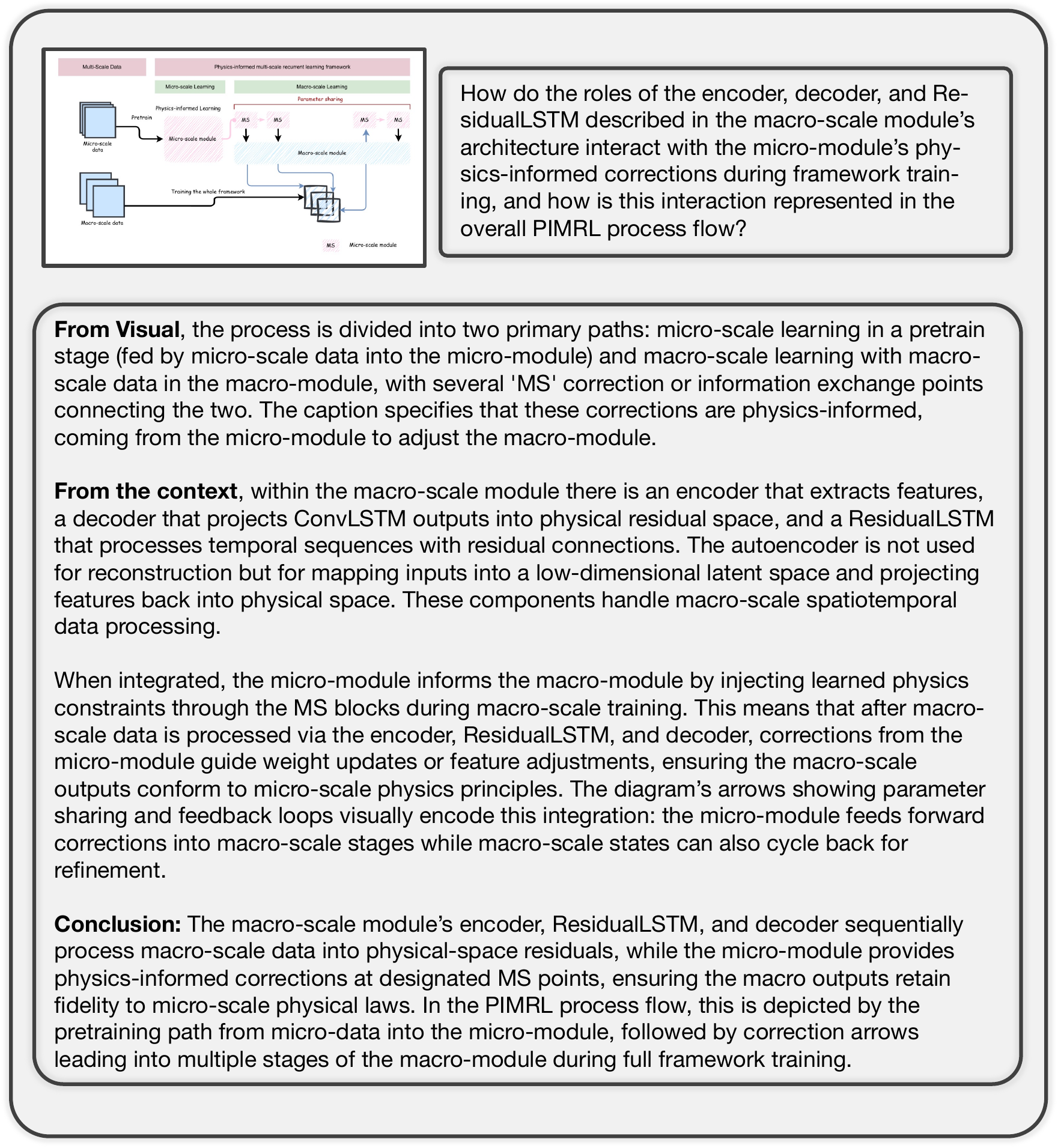}
\caption{\textbf{Example of CIM (Concept-to-Instance Mapping) type question.} This example shows how the model links abstract architectural components (encoder, decoder, ResidualLSTM) described in text to their concrete visual representations in the system diagram, tracing information flow across modules.}
\label{fig:example_cim}
\end{figure*}

\paragraph{Concept-to-Instance Mapping (CIM).} \autoref{fig:example_cim} shows a CIM-type question that links abstract architectural concepts described in text to their concrete visual representations in diagrams.

\begin{figure*}[t]
\centering
\includegraphics[width=0.95\linewidth]{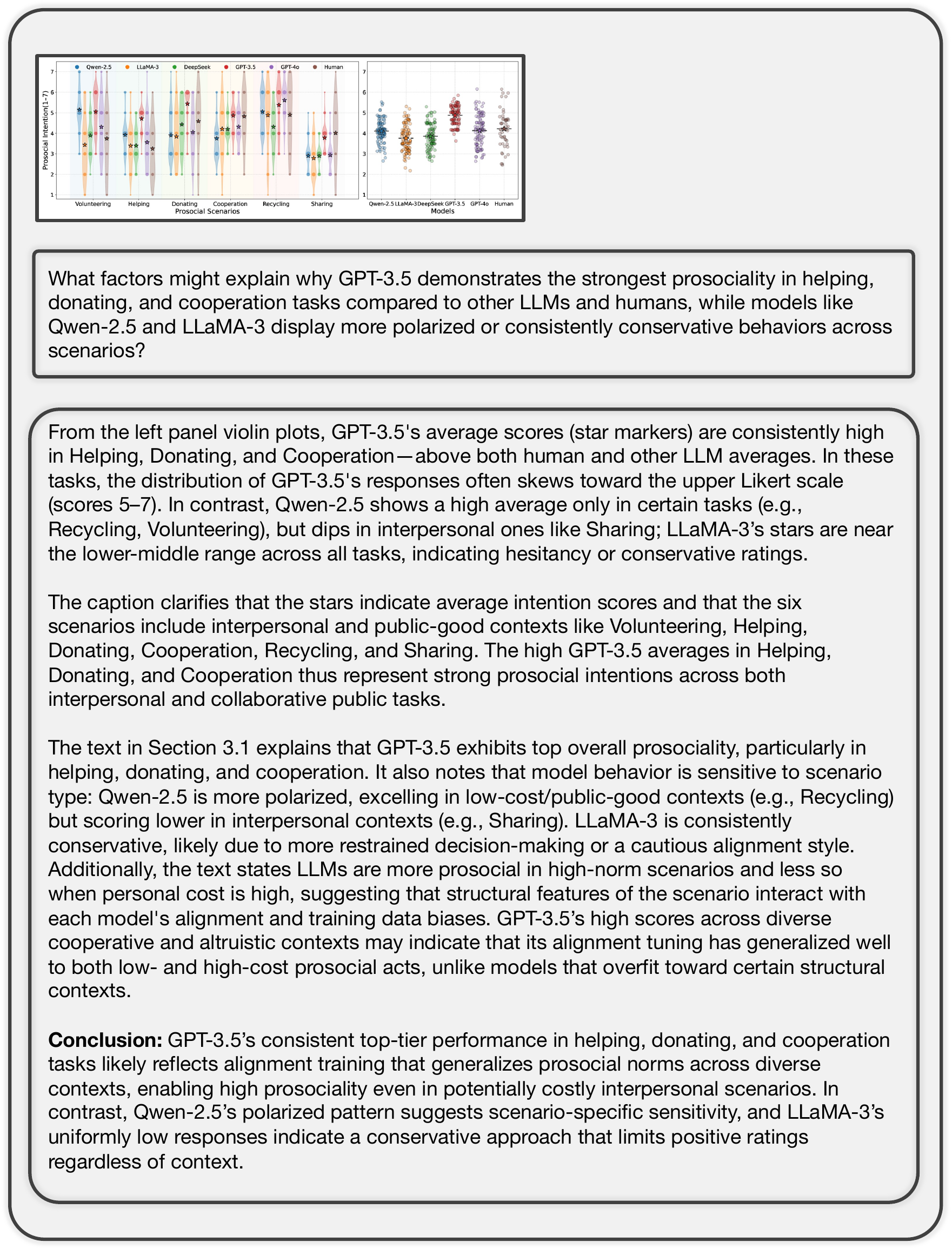}
\caption{\textbf{Example of HVI (Hypothesis Validation \& Inferential Reasoning) type question.} This example illustrates inferential reasoning where the model analyzes distributional patterns in violin plots alongside textual explanations to infer underlying factors explaining behavioral differences across models.}
\label{fig:example_hvi}
\end{figure*}

\paragraph{Hypothesis Validation \& Inferential Reasoning (HVI).} \autoref{fig:example_hvi} presents an HVI-type question demonstrating inferential reasoning by synthesizing visual patterns and textual explanations to draw conclusions.

\begin{figure*}[t]
\centering
\includegraphics[width=0.95\linewidth]{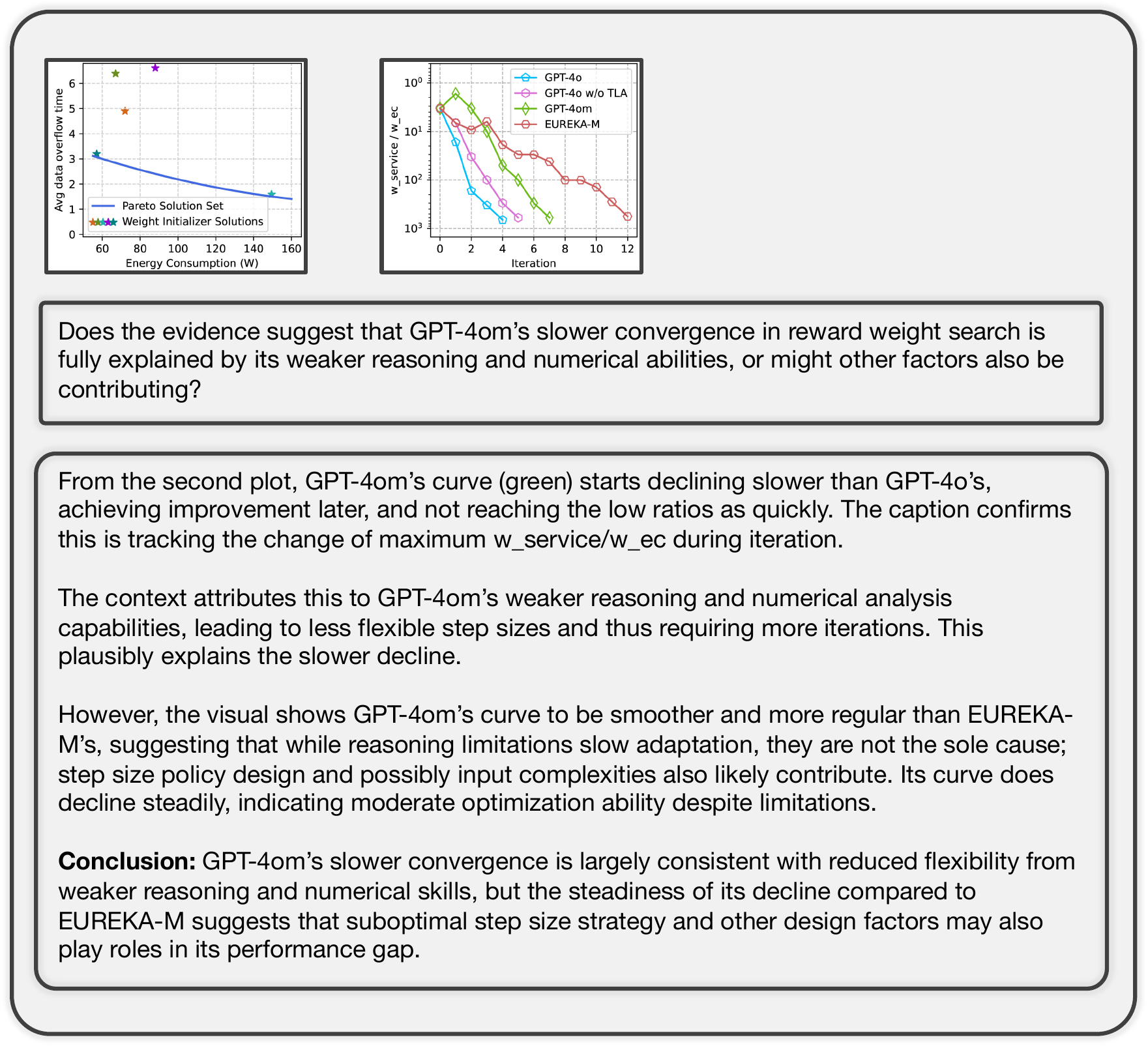}
\caption{\textbf{Example of CAC (Critical Analysis \& Consistency Check) type question.} This example demonstrates critical evaluation of whether textual claims are accurately supported by visual data, requiring careful assessment of evidence strength and potential discrepancies.}
\label{fig:example_cac}
\end{figure*}

\paragraph{Critical Analysis \& Consistency Check (CAC).} \autoref{fig:example_cac} provides a CAC-type question that critically evaluates the consistency between textual characterizations and visual data.

\begin{figure*}[t]
\centering
\includegraphics[width=0.95\linewidth]{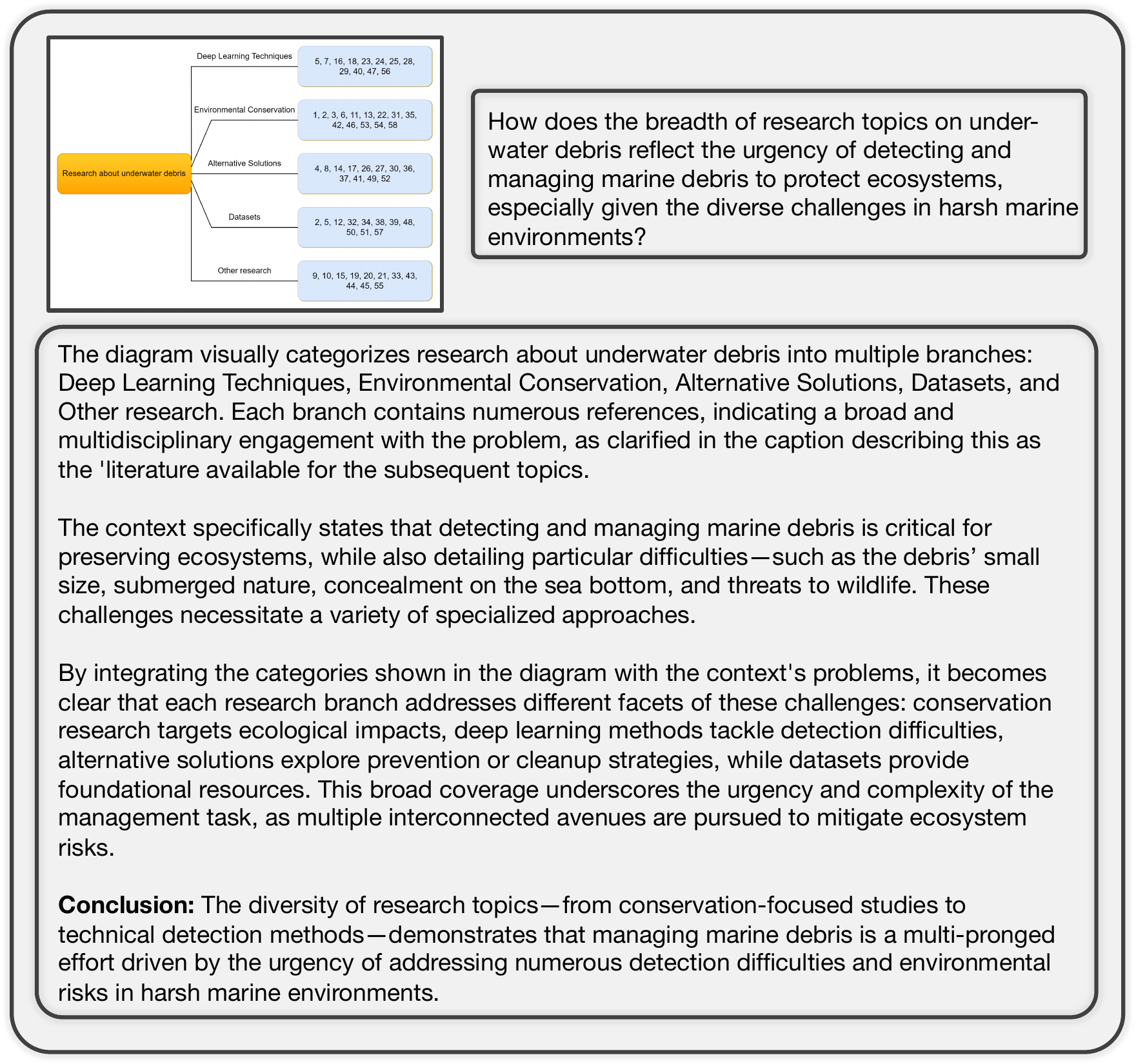}
\caption{\textbf{Example of ARS (Argumentative Role \& Synthesis) type question.} This example shows how the model synthesizes visual evidence and textual arguments to articulate the overall scientific contribution and understand the role of visual elements in supporting the main thesis.}
\label{fig:example_ars}
\end{figure*}

\paragraph{Argumentative Role \& Synthesis (ARS).} \autoref{fig:example_ars} displays an ARS-type question requiring synthesis of visual evidence and textual arguments to understand the overall scientific contribution.

\end{document}